%% file: 00_main_abstract.tex
\documentclass[5p,12pt]{elsarticle}

\usepackage{fancyhdr}

\usepackage{color}
\usepackage[cmex10]{amsmath}
\usepackage{bm}
\usepackage{amssymb}
\usepackage{multirow}
\usepackage[ruled, lined, linesnumbered, commentsnumbered]{algorithm2e}  
\usepackage{caption}
\usepackage{subcaption}
\usepackage{tikz}
\usepackage{eurosym}
\usepackage{stmaryrd}
\usepackage{hyperref}

\usepackage{nomencl}
\makenomenclature

\RequirePackage[normalem]{ulem} 
\RequirePackage{color}\definecolor{RED}{rgb}{1,0,0}\definecolor{BLUE}{rgb}{0,0,1} 

\include{NN_MCMC/precomands}
\include{NN_MCMC/nomeclature.tex}

\journal{Applied Energy}

\begin{document}

\begin{frontmatter}

\title{Deep Learning-enabled MCMC for Probabilistic State Estimation in District Heating Grids
}

\author{Andreas Bott, Tim Janke, Florian Steinke}
\address{Energy Information Networks \& Systems, Technical University of Darmstadt, Landgraf-Georg-Str. 4, 64283 Darmstadt, Germany}




\begin{abstract}
Flexible district heating grids form an important part of future, low-carbon energy systems.
We examine probabilistic state estimation in such grids, 
i.e., we aim to estimate the posterior probability distribution over all grid state variables such as pressures, temperatures, and mass flows
conditional on measurements of a subset of these states.
Since the posterior state distribution does not belong to a standard class of probability distributions, we use Markov Chain Monte Carlo (MCMC) sampling in the space of network heat exchanges and evaluate the samples in the grid state space to estimate the posterior.
Converting the heat exchange samples into grid states by solving the non-linear grid equations makes this approach computationally burdensome. 
However, we propose to speed it up by employing a deep neural network that is trained to approximate the solution of the exact but slow non-linear solver.
This novel approach is shown to deliver highly accurate posterior distributions
both for classic tree-shaped as well as meshed heating grids,
at significantly reduced computational costs that are acceptable for online control.
Our state estimation approach thus enables tightening the safety margins for temperature and pressure control and thereby a more efficient grid operation.
\end{abstract}

\begin{keyword}
state estimation, district heating grids, probabilistic state estimation, deep neural networks, Markov Chain Monte Carlo 
\newline
\newline
\copyright 2023 This manuscript version is made available under the CC-BY-NC-ND 4.0 license \href{https://creativecommons.org/licenses/by-nc-nd/4.0/}{https://creativecommons.org/licenses/by-nc-nd/4.0/} 
\end{keyword}

\end{frontmatter}
\printnomenclature
\input{NN_MCMC/02_paper}

\bibliography{NN_MCMC/bibfile}
\input{NN_MCMC/10_appendix}

\end{document}

%% file: NN_MCMC/precomands.tex
\newcommand{\vh}{\mathbf{h}}

\newcommand{\vm}{\mathbf{m}}

\newcommand{\vq}{\mathbf{q}}

\newcommand{\vx}{\mathbf{x}}

\newcommand{\mxj}{\mathbf{J}}

\newcommand{\tamb}{T_{amb}}
\newcommand{\Grid}{G}
\newcommand{\Nodes}{V}
\newcommand{\Neighbours}[1]{N_{#1}}
\newcommand{\Edges}{E}
\newcommand{\temp}[1]{T_{#1}}
\newcommand{\pr}[1]{p_{#1}}
\newcommand{\tempend}[1]{\temp{#1}^{end}}
\newcommand{\mf}[1]{\dot{m}_{#1}}
\newcommand{\demand}[1]{q_{#1}}
\newcommand{\vdemand}{\vq}
\newcommand{\state}{\vx}
\newcommand{\NNstate}{\hat{\state}}
\newcommand{\setpoints}{\bm{\eta}}
\newcommand{\NNweight}{\bm{\theta}}
\newcommand{\ND}{\bm{N}}
\newcommand{\prob}{p}

\newcommand{\vmval}{\vm}
\newcommand{\mCov}{\Sigma_{m}}

\newcommand{\demandMean}{\vmu_{\vdemand}}
\newcommand{\demandCov}{\Sigma_{\vdemand}}

\newcommand{\statelin}{{LSE}}
\newcommand{\Jacobian}[1]{\mxj_{#1}}

\newcommand{\statelinCov}{\Sigma_{\statelin}}

\newcommand{\indexMatrix}[1]{\Xi_{#1}}

\newcommand{\statequations}{\mathbf{e}}
\newcommand{\modelDS}{\mathbf{h}}
\newcommand{\NN}{\hat{\vh}_{\NNweight}}

\newcommand{\PTGrid}{grid-loop}
\newcommand{\RealGrid}{grid-tree}

\newcommand{\seLoss}{\psi}

\newcommand{\vmu}{\mathbf{\mu}}
\newcommand{\Expval}{\mathbb{E}}

\newcommand{\fig}[1]{Figure~\ref{fig:#1}}
\newcommand{\tab}[1]{Table~\ref{tab:#1}}

%% file: NN_MCMC/nomeclature.tex
\usepackage{etoolbox}
\renewcommand\nomgroup[1]{%
  \item[\bfseries
  \ifstrequal{#1}{A}{General Abbreviations}{%
  \ifstrequal{#1}{G}{Grid Parameter}{%
  \ifstrequal{#1}{H}{Grid State Variables}{%
  \ifstrequal{#1}{M}{Probabilistic State Description}{%
  }}}}%
]}

\nomenclature[A, 03]{MCMC}{Markov Chain Monte Carlo}
\nomenclature[A, 04]{DNN}{Deep Neural Networks}
\nomenclature[A, 01]{NR}{Newton Raphson Algorithm}
\nomenclature[A, 02]{SIR-MC}{Sample importance resampling \\Monte Carlo}
\nomenclature[A, 06]{LSE}{Linarisation-based probabilistic \\state estimation}
\nomenclature[A, 05]{MCMC-DNN}{Proposed framework for \\state estimation}

\nomenclature[G, 01]{$\Grid$}{Graph describing the district heating grid $G = (\Nodes ,\Edges)$}
\nomenclature[G, 02]{$\Nodes$}{Nodes of the graph $\Grid$}
\nomenclature[G, 03]{$\Edges$}{Edges of the graph $\Grid$; $\Edges = \Edges^{active} \cup \Edges^{passive}$}
\nomenclature[G, 04]{$\Edges^{passive}$}{Passive edge, i.e. pipe}
\nomenclature[G, 06]{$\Edges^{demand}$}{Heat consumer edge}
\nomenclature[G, 07]{$\Edges^{source}$}{Heat producer edge}
\nomenclature[G, 08]{$\Edges^{slack}$}{Slack edge (heat producer supplying \\balancing power)}
\nomenclature[G, 05]{$\Edges^{active}$}{Active grid edges; \\$\Edges^{active} = \Edges^{demand} \cup \Edges^{source} \cup \Edges^{slack}$}
\nomenclature[G, 09]{$\Neighbours{i}$}{Set of neighbours of node $i$}
\nomenclature[G, 10]{$l_{ij}$}{Length of pipe $(i,j)$}
\nomenclature[G, 11]{$k_{ij}$}{Specific pressure loss per pipe length for edge $(i,j)$}
\nomenclature[G, 12]{$\lambda_{ij}$}{Specific heat loss per pipe length for edge $(i,j)$}

\nomenclature[G, 13]{$c_p$}{Specific heat capacity of heating fluid}

\nomenclature[H, 02]{$\temp{i}$}{Temperature at node $i$}
\nomenclature[H, 03]{$\pr{i}$}{Pressure at node $i$}
\nomenclature[H, 04]{$\mf{ij}$}{Mass flow through edge $(i,j)$}
\nomenclature[H, 05]{$\tempend{ij}$}{Temperature at the outlet of edge $(i,j)$}
\nomenclature[H, 09]{$\temp{ij}^{set}$}{Temperature setpoint for the outlet of active edge $(i,j)$}
\nomenclature[H, 10]{$\pr{i}^{set}$}{Pressure setpoint for node $i$}
\nomenclature[H, 01]{$\state$}{Grid state vector; \\${\state := [(\temp{i}, \pr{i})_{i\in\Nodes}, (\mf{ij}, \tempend{ij})_{(i,j) \in \Edges}]}$}
\nomenclature[H, 06]{$\demand{ij}$}{Heat power consumed/produced at edge $(i,j)$}
\nomenclature[H, 07]{$\vdemand$}{Vector containing the heat powers of all active edges}
\nomenclature[H, 08]{$\setpoints$}{Vector containing all grid control parameters; \\$\setpoints := [(\temp{ij}^{set})_{(i,j) \in \Edges^{active}},(\pr{i}^{set}, \pr{j}^{set})_{(i,j) \in \Edges^{slack}}]$}
\nomenclature[H, 12]{$\tamb$}{Ambient temperature}
\nomenclature[H, 11]{$\statequations$}{Operator summarizing grid equations; \\$\statequations \left(\state, \vdemand, \setpoints \right) = 0$}

\nomenclature[M, 01]{$\prob(\cdot)$}{Probability density function}
\nomenclature[M, 05]{$\vmval$}{Vector containing all measured values}
\nomenclature[M, 02]{$\ND (\vmu , \Sigma )$}{Multivariate normal distribution with mean $\vmu$ and covariance matrix $\Sigma$}
\nomenclature[M, 03]{$\ND^+(\mu, \Sigma$)}{Zero truncated normal distribution}
\nomenclature[M, 06]{$\indexMatrix{M}$}{Binary matrix selecting the measured states from the full state vector $\state$}
\nomenclature[M, 04]{$\delta (\cdot)$}{Dirac delta function}
\nomenclature[M, 12]{$\omega_i$}{Weight associated with sample $i$}
\nomenclature[M, 13]{$g(\cdot)$}{Markov process generating candidate samples during MCMC procedure}

\nomenclature[M, 07]{$\modelDS$}{Mapping between the heat exchanges vector $\vdemand$ and the grid state $\state$; $\state = \modelDS(\vdemand)$}
\nomenclature[M, 08]{$\NN$}{Neural network approximating $\modelDS$ with parameters $\NNweight$}
\nomenclature[M, 09]{$\NNstate$}{State predicted by the neural network; $\NNstate = \NN(\vdemand)$}
\nomenclature[M, 10]{$\Jacobian{}$}{Jacobian matrix of mapping $\modelDS$; $\Jacobian{} = \partial \modelDS / \partial \vdemand$}
\nomenclature[M, 17]{$\kappa$}{Index indicating the type of physical quantity; $\kappa \in \{\temp{}, \pr{}, \mf{}, \tempend{}\}$}
\nomenclature[M, 16]{$\seLoss (\cdot )$}{Mean squared violation of state equations; $\seLoss = \frac{1}{N} \sum_i \lVert \statequations(\NNstate_i, \vdemand_i, \setpoints) \rVert ^2$}
\nomenclature[M, 15]{$\mathcal{E}_{\alpha}(X, Y)$}{Energy distance between random variables $X$ and $Y$}
\nomenclature[M, 18]{$q_5(X)$}{5\% quantile of random variable $X$}
\nomenclature[M, 19]{$\Delta q5$}{Mean absolute error of the predicted 5\% quantile}
\nomenclature[M, 20]{$\Delta m$}{Mean absolute error of the predicted mean}
\nomenclature[M, 14]{$\Expval (\cdot)$}{Expected value}

%% file: NN_MCMC/02_paper.tex
\newpage
\section{Introduction} \label{introduction}
\subsection{Motivation}
The electrification and flexibilisation of heating is required both for decarbonising the heating sector and the electric power sector \cite{palzer2014comprehensive, thomassen2021decarbonisation}. 
Coupling electricity and heat is beneficial for both sectors as it can help to reduce the CO2 emissions in the heating sector and, at the same time, can provide flexibility to balance variable renewable energy supplies in the electricity sector.
A key enabler for the coupling are 4th generation district heating grids \cite{lund20144th,lund2018status}.
They are characterised by lower supply and return temperatures and a more flexible grid operation compared to traditional district heating grids. Lower temperatures enable a more efficient use of power to heat technologies such as heat pumps
and the integration of industrial waste heat.
Storages and consumer flexibility allow to utilise of fluctuating heat sources \cite{prina2016smart, mathiesen2015ida}. 
The topology of the heating grids changes as well. The traditionally tree-based design shifts towards more loop-based designs to facilitate the incorporation of distributed heat sources \cite{Flexynets2016}.

In this context, state estimation for heating grids becomes increasingly important, to enable ever more flexible operation schemes than used today \cite{vandermeulen2018controlling, novitsky2020smarter}. 
The ultimate goal of grid operation is to ensure acceptable supply conditions for the customers, i.e., sufficiently high temperatures and pressure differences from the supply to the return system, while not generating unnecessary losses due to too high temperatures and pressures.
The temperatures and pressures in the grid depend on grid operators' pumps and the heat supply conditions, but also on the consumers' behaviour which is typically unknown to the operator.

In a traditional, tree-like grid layout with only one heat source, pressures and temperatures monotonically decrease with the distance to the heat plant. 
Measurements at the point furthest away from the heat plant are thus classically used to adjust the supply and pumping conditions.
However, in grids with loops or decentralised feed-ins, it is not a priori evident where to measure the lowest supply pressure and temperature.

If not all locations are to be measured and excessive safety margins for temperatures and pressures are to be avoided, 
a reliable and precise estimation of the grid's state, i.e., the temperatures and pressures at all locations, is needed.
This estimate should include uncertainty intervals to account for the customers' behavior that is unknown to the grid operator. 
One option to achieve this is probabilistic state estimation,
computing the probability distribution over the states conditioned on all knowledge available to the grid operator, the so-called posterior distribution.
As we will show in the experiment section, for heating grids these distributions do not belong to any standard class of probability distributions. They can be highly skewed or even multimodal. 
Thus merely estimating the most probable or average state value might be highly misleading. 

This work, therefore, proposes a new probabilistic state estimation approach which is applicable to modern heating networks including loops, does not need any prior assumptions regarding the form of the state distributions, and reaches calculation times in the range needed for online decision-making in heating grids.

\subsection{Literature Review}
State estimation in heating grids is conceptually related to state estimation in electric power systems 
which has received significant attention
\cite{primadianto2016review}.
In this realm, machine learning and deep learning have been applied in various ways.
Deep Neural Network (DNN) predictions of the network state given measurement values can be used to kick-start a Newton Raphson (NR) solver for deterministic state estimation \cite{zamzam2019data}.
The conditional mean state estimates, as opposed to the maximum likelihood estimates, can also be  computed with a learned model \cite{mestav2019bayesian}.
DNNs can be trained in a physics-aware fashion \cite{zamzam2020physics} 
or by exploiting the graph structure via graph neural networks to predict power flow \cite{donon2020neural}.
To determine full posterior state distributions, one can use a combination of DNNs and Gaussian mixture models
tailored to electric grid properties \cite{huang2019probabilistic}
or use a linearisation approach \cite{schenato2014bayesian}.

For district heating grids the topic of state estimation has received far less attention. 
Existing approaches can be separated into two groups depending on whether they consider time delays in the network. 
Analyses including time delays are often used in the context of CHP operation optimisation, e.g., in \cite{sheng2017two, zhang2019decentralized}. Both approaches only consider small radial district heating grids and don't provide uncertainties for the estimated states. 
A probabilistic approach using time-dependent models and Gaussian Processes \cite{simonson2020probabilistic} 
yields good results for the uncertainty modelling of a single pipe but reports problems in terms of scalability to larger networks. It also does not consider bidirectional mass flows. 
For not too large grids or minor changes in the supply temperatures and heat exchanges, the dynamic system state will deviate only little from the steady state. 
This, together with the tremendous simplification of the modelling and faster computation, renders steady state modelling popular for district heating grids. 
Previous work has considered deterministic state estimation with complete load measurements \cite{fang2014state}
or with only partial information \cite{zhang2021state}. 
One route to probabilistic steady-state estimation in heating grids is linearising the grid equations around the best state estimate and using standard propagation rules for normal distributions to determine Gaussian posterior distributions \cite{BotSte21, MatBotReh21}.
For radial networks, non-linear approximation of the grid equations can be used as well to determine Gaussian posterior representations \cite{sun2019nonlinear}.

However, the true posterior distributions often do not fall into any standard class of distributions but are often highly skewed to one side \cite{BotSte21}. And, as our experiments show, these distributions may even be multimodal, especially for grids featuring loops. 
Robust estimates of pressures and temperatures based on quantiles of such fixed distribution shapes or safety margins based on deterministic state estimates may thus be highly misleading.

\begin{figure}[t]
     \centering
     \begin{subfigure}[b]{0.4\textwidth}
         \centering
         \includegraphics[width=\linewidth]{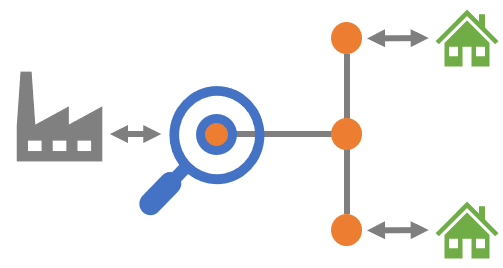}
        \caption{Physical district heating grid}
         \label{fig:fig1a}
     \end{subfigure}
     \begin{subfigure}[b]{0.4\textwidth}
         \centering
         \includegraphics[width=\linewidth]{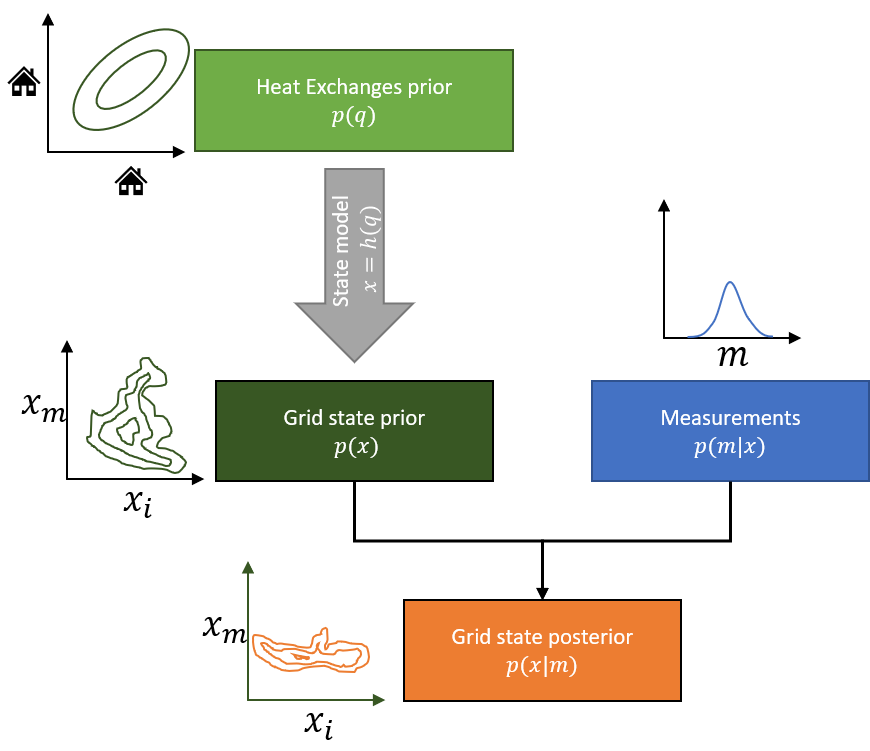}
        \caption{Sketch of the mathematical approach}
         \label{fig:fig1b}
     \end{subfigure}
        \caption{
        (a) Given measurements at a few nodes (blue), we aim to estimate the grid state at the other grid nodes (orange).
        (b) The prior distribution over grid states (dark green) is not in a class of standard distributions and is thus determined indirectly via the distribution of heat exchanges (light green).
        Bayes rule allows us to derive the state posterior (orange) given a measurement likelihood model (blue).
        }
        \label{fig:fig1}
\end{figure}

\subsection{Our Contribution}
This paper extends the existing work on probabilistic state estimation within the setting of steady-state modelling of district heating grids. The main benefit of our new, more versatile approach is, that it doesn't make any assumption regarding a specific form of the posterior distribution.
Specifically, we assume the actual heat consumption at each demand to be uncertain,
while some grid states, such as temperatures, pressures, or mass flows, are measured. The measurements are corrupted by measurement noise and the set of measurements is incomplete, i.e., the measurements do not uniquely determine the actual grid state, a situation that is practically very common. 
We then estimate conditional probability distributions over all grid states given these measurements. 
\fig{fig1} sketches out how the uncertain quantities are connected in our uncertainty model. 

Given a prior distribution over states and a measurement likelihood, Markov Chain Monte Carlo (MCMC) algorithms create Markov chains that have an equilibrium distribution proportional to the desired posterior distribution.
While we are ultimately interested in the posterior over the grid states, 
running Markov chains in the grid state space is impractical.
This is because randomly generated proposals for state vector updates are very unlikely to be consistent with the physical equations of the heating grid.
Instead, we run the Markov chains in the space of the heat exchanges, i.e., the thermal demands and heat sources. Here, a prior distribution can more naturally be constructed either from historical data or using load profile modelling.
Note also that in this space no dependencies between the different dimensions are implied via the grid equations. 

Sampling heat exchanges allows us to indirectly sample from the desired distribution over the grid states.
This is because each combination of heat exchange values physically corresponds to a unique network state, given defined outlet temperatures at the heat exchanges.
The mapping from the heat exchanges to grid states thus is unique.
Updating the Markov chain state requires evaluating each proposal regarding its likelihood which is computationally costly as it is defined in the state space and therefore requires solving the non-linear state equations for every newly proposed update.
Computing the grid state via a non-linear equation solver such as NR renders this approach computationally prohibitive for online operation \cite{tol2021development}.

Instead, we propose to use a DNN that is trained to approximate the solution of the NR solver to speed this step up by several orders of magnitude without significantly reducing the quality of the solution.
The DNN also enables the use of the efficient derivative-based Hamiltonian MCMC algorithm.


The presented approach is demonstrated on both a traditional tree-like district heating grid and a grid topology featuring a loop structure. 
We show strongly superior approximation properties for the true posterior distributions compared to the linearisation approach \cite{MatBotReh21,BotSte21}.
At the same time, computation times 
are several orders of magnitude lower compared to solving the state equations via NR for each sample of a sample importance resampling Monte Carlo (SIR-MC) approach \cite{rubin1987calculation},
which we use as a baseline since MCMC with NR is not computationally feasible.
Overall, we achieve computational times in the range of a few minutes, which is suitable for online heating grid control.
Our contribution is thus an important step towards enabling smart, flexible, low-carbon district heating systems.

The remainder of this paper is structured as follows. The steady-state heat grid equations that are the basis of this work are presented in section \ref{sec:model}.
Section \ref{sec:problem} introduces the topic of state estimation from a probabilistic standpoint  and presents our proposed approach.
Details on the implementation can be found in section \ref{sec:implementation}, while the results of the simulation experiment are discussed in section \ref{sec:results}. We conclude in section \ref{sec:conclusion}.


\section{Steady-State Heat Grid Model} \label{sec:model}
We model the heating network as a graph $ \Grid = (\Nodes , \Edges)$ with nodes $\Nodes$ and edges $\Edges \subseteq \Nodes \times \Nodes$.
For each node $i \in \Nodes$ we denote the heating fluid temperature by $\temp{i}$ and its pressure by $\pr{i}$.
For edge $(i, j) \in \Edges$, let $\mf{ij}$ denote the mass flow rate from $i$ to $j$ and $\tempend{ij}$ the fluid temperature at the outlet of the edge.
Symmetry implies $\mf{ij} = - \mf{ji}$ for all edges $(i,j) \in \Edges$. 
We further denote the set of neighbours of node $i$ by $\Neighbours{i}$.
The heat exchange with the external world over edge $(i,j)\in\Edges$ is denoted by $\demand{ij}$, where positive values represent heat consumption. 

We assume steady-state conditions and neglect time delays in the network. 
The heating fluid, typically hot water, is assumed to be incompressible and to have constant fluid properties. 
The heat grid is then described by the following thermal and hydraulic equations.

Conservation of mass dictates
\begin{align}
    &\sum_{j\in\Neighbours{i}} \mf{ij} = 0, & \forall i \in \Nodes. \label{eq_mconv}
\end{align}
Assuming perfect mixing, energy conservation mandates 
\begin{align}
    \temp{i} = \frac{\sum_{j\in \Neighbours{i}, \mf{ji} > 0}\mf{ji} \tempend{ji}}{\sum_{j\in \Neighbours{i}, \mf{ij} > 0}\mf{ij}}, \quad \forall  i \in \Nodes.  \label{eq:tmix}
\end{align}
We distinguish between passive and active edges, i.e.,  ${\Edges = \Edges^{passive} \cup \Edges^{active}}$. 
Passive edges are pipes that are only subject to physically determined heat and pressure losses.
They are characterised by 
\begin{align}
    \pr{i} - \pr{j} = {} & k_{ij} \mf{ij} \left| \mf{ij} \right|, \notag \\ & \forall (i,j) \in \Edges^{passive}, \label{eq:pipeP} \\
    \tempend{ij} &= \left(\temp{i} - \temp{a}\right) \text{exp} \left(-\frac{l_{ij}\lambda_{ij}}{c_p \mf{ij}}\right) + \temp{a}, \notag \\  & \forall (i,j) \in \Edges^{passive},  \label{eq_pipeT}
\end{align}
where $c_p$ denotes the specific heat capacity, $l_{ij}$ is the length of the pipe, $\temp{a}$ the pipe ambient temperature, and $k_{ij}$ and $\lambda_{ij}$ are parameters that characterise the specific pressure or heat losses of the pipe.
We assume these parameters to be constant within our temperature and mass flow range, an assumption validated in \cite{BotSte21}.

Active edges represent heat sources or consumers whose behaviour is defined by an external control strategy.
Consumer edges consist of a heat exchanger and its local controller. 
The controller adjusts the heating fluid flow such that the necessary heat $\demand{ij}$ is exchanged and the temperature at the end of the heat exchanger is close to a setpoint $\temp{ij}^{set}$, which is typically specified by the grid operator.
Assuming that the heat exchanger and controller work as specified, we model 
\begin{align}
    &\tempend{ij} = \temp{ij}^{set}, & \forall (i,j) \in \Edges^{demand}, \label{eq:fix1}\\
    &\demand{ij} = \mf{ij}c_p \left(\temp{i}-\tempend{ij}\right), & \forall (i,j) \in \Edges^{demand}.\label{eq:qDemand}
\end{align}
The pressure drop over active edges is not defined explicitly, as it is adjusted locally via valves to ensure the desired mass flow. 
Heat sources $\Edges^{source}$ with a fixed heat output are modelled analogously, transferring fluid from the return side to the supply side of the grid. 

The modelling equations above determine the pressures in the grid only up to a constant offset for the supply side and the return side, respectively. These pressure levels are controlled by central pumps to ensure a sufficient pressure drop at each consumer and to avoid evaporation in grids with superheated water. 
Additionally, in a steady-state the total heat supply has to match the total consumption and losses in the grid. Both requirements are typically ensured by one slack generator $(i,j) \in \Edges^{slack}$ which is modelled as
\begin{align}
    & \tempend{ij} = \temp{ij}^{set}, & \forall (i,j) \in \Edges^{slack},\label{eq:fix2}\\ 
    & \pr{i} = \pr{i}^{set}, \pr{j} = \pr{j}^{set}, & \forall (i,j) \in \Edges^{slack}. \label{eq:fix3}
\end{align}
Note that the choice of nodes for fixing the absolute pressure levels is arbitrary in this model, as long as one is part of the supply side and the return side each. 
We then have $\Edges^{active} = \Edges^{source} \cup \Edges^{demand} \cup \Edges^{slack}$.

\bigskip
Equations \eqref{eq_mconv} - \eqref{eq_pipeT} form a system of non-linear equations that can be compactly denoted as follows.
The state $\state$ of the system can be summarised as \\ ${\state := [(\temp{i}, \pr{i})_{i\in\Nodes}, (\mf{ij}, \tempend{ij})_{(i,j) \in \Edges}]}$,
the heat exchanges as \\${\vdemand := [(\demand{ij})_{(i,j) \in \Edges^{source}\cup\Edges^{demand}}]}$,
and the control parameters as \\${\setpoints := [(\temp{ij}^{set})_{(i,j) \in \Edges^{active}},(\pr{i}^{set}, \pr{j}^{set})_{(i,j) \in \Edges^{slack}}]}$.
Equations \eqref{eq_mconv} - \eqref{eq:fix3} can then be denoted jointly as the state equation
\begin{align} \label{eq:state}
    \statequations \left(\state, \vdemand, \setpoints \right) = 0
\end{align}
with a suitably defined non-linear operator $\statequations$.
The control parameters $\setpoints$ are assumed to be known and fixed throughout this paper.
The implicit function theorem then states that there exists a mapping
\begin{align} \label{eq:model-ds}
    \state = \modelDS(\vdemand)
\end{align}
between the network's state and the corresponding heat exchange values at least locally around every $(\vdemand, \state)$ pair, that fulfils \eqref{eq:state} \cite{hoehre_math_fuer_ingeneure}. 
However, this mapping has no closed form representation. 
Therefore, solving \eqref{eq:model-ds} for any given input $\vdemand$ requires implicitly calling a non-linear system solver for \eqref{eq:state} such as the NR algorithm.
We will propose below to approximate this process with a DNN.

\section{Probabilistic State Estimation} \label{sec:problem}
The model described above allows for determining the grid state $\state$ if all heat exchanges $\vdemand$ are known. However, this information often is not fully available when grid control decisions have to be taken.
Under the assumption of incomplete information, i.e., measurements that do not uniquely determine the grid state, probabilistic state estimation aims at estimating the distribution over all states, given all available information. 

Specifically, we assume to have measurements $\vmval$ for a subset of the grid states, e.g., of Temperatures, pressures, and mass flows at some location. 
Our goal is then to obtain the posterior distribution over all other states, i.e., estimate the distribution $p(\state|\vmval)$.
The choice, of which state variables are measured, is arbitrary to the approach described below, as long as at least one state variable is measured and its value is not a priori fixed via equation \eqref{eq:fix1}, \eqref{eq:fix2} or \eqref{eq:fix3}. Of course, the informativeness of the selected measurements will influence the variance of the resulting probability distributions. 

\subsection{Bayes' Theorem for state estimation}\label{sec:bayes}
Bayes' theorem states that 
\begin{equation} \label{eq:bayes}
    p(\state|\vmval) = \frac{p(\vmval|\state) p(\state)}{p(\vmval)} \propto p(\vmval|\state) p(\state),
\end{equation}
where $p(\state|\vmval)$ is called the posterior, $p(\vmval|\state)$ is the likelihood,  $p(\state)$ the prior.
$p(\vmval)$ is a normalising constant
that is not needed for sampling-based approaches.
For the likelihood we assume throughout that the measurements are corrupted by independent Gaussian noise, i.e.,  $p(\vmval|\state) = N(\vmu_m, \mCov)$, where $N$ denotes the normal distribution. Its mean is given by $\vmu_m = \indexMatrix{M} \state$ where $\indexMatrix{M}$ is a binary matrix that selects the measured states. The covariance matrix $\mCov$ is a diagonal matrix encoding the measurement uncertainties. 

A classical Monte Carlo (MC) approximation for the posterior $\prob(\state|\vmval)$ distribution would sample $N$ independent and identically distributed (iid) states $\{\state_i\}_{i=1}^N$ and weight each sample according to its likelihood $p(\vmval|\state_i)$. 
However, drawing random samples directly in state space is practically not possible, 
as only physically feasible system states $\state_i$ which fulfil the modelling equations \eqref{eq_mconv} - \eqref{eq:fix1} and \eqref{eq:fix2} - \eqref{eq:fix3} have a nonzero probability. 
The possible states thus lie on a submanifold of the state space and cannot be described by any standard class of distributions in this space.
Drawing state samples indirectly via rejection sampling, i.e., sampling states from some other distribution and discarding all samples that are not physically feasible, is highly inefficient as well as nearly all samples would be discarded. 
We thus propose a different approach.

\subsection{MC sampling in the heat exchange space} \label{sec:MC}
Each feasible state uniquely corresponds to some heat exchange $\vdemand$ via \eqref{eq:qDemand}. 
This allows us to use the sifting property of the Dirac delta function $\delta (\cdot )$ together with the mapping \eqref{eq:model-ds} to express the prior over the states $p(\state)$ in terms of a prior over the heat exchanges $p(\vdemand)$ as
\begin{equation} \label{eq:poststateprob}
    p(\state) 
    = \int 
    p(\vdemand)
    \delta{(\state - \modelDS(\vdemand))}  d\vdemand.
\end{equation}
We can then rewrite the unnormalised posterior \eqref{eq:bayes} to express it in terms of heat exchanges as 
\begin{align}
    p(\vmval|\state) p(\state) =& \int
    p(\vmval|\state) 
    \delta{(\state - \modelDS(\vdemand))} p(\vdemand) d\vdemand \nonumber\\
    =& p(\vmval|\modelDS(\vdemand)) p(\vdemand).
\end{align}

Defining a valid prior over the heat exchanges is possible, as the entries of these vectors are not physically interdependent. The prior could be based on historical data, probabilistic heat demand forecasting or production schedules. 
In our case, we model the prior of the demands as zero truncated normal distribution 
\begin{align}
p(\vdemand) = \ND^+(\demandMean, \demandCov),
\end{align}
with the mean vector $\demandMean$ and covariance matrix $\demandCov$.
The probability density function of the zero-truncated normal distribution is exactly zero if any entry of $\vdemand$ is negative. 
Otherwise, it is proportional to a normal distribution with the same parameters. 
The truncation is done to exclude unrealistic negative demands from the distribution. 
For feed-ins, the absolute value of the heat exchange would be sampled from this distribution and negated afterwards.

The points discussed above give rise to the Monte Carlo Sampling Importance Resampling (MC-SIR) algorithm as the first algorithm for probabilistic state estimation.
It is summarised in Algorithm \ref{alg:MC}. 
First, $N$ iid samples $\{\vdemand_i\}_{i=1}^N$  are drawn from the heat exchange prior. Then \eqref{eq:model-ds} is solved for each sample using the NR algorithm to obtain state samples $\{\state_i\}_{i=1}^N$. 
Using the given measurement $\vmval$ each sample is weighted with its likelihood $\omega_i = \prob\left(\vmval|\state_i\right)$.
Most MC samples have a very small weight assigned and therefore contribute little to the estimated distribution. We use a resampling step to represent the posterior distribution with a strongly reduced number of samples.  
Therefore, samples are drawn with replacement from the initial MC samples $\{\state_i\}_{i=1}^N$, where the probability of drawing any specific sample $j$ is given by its normalised weight $p(\state_j) = \omega_j / \sum_i \omega_i$. This approach is known as a Sampling Importance Resampling \cite{rubin1987calculation}. 

\begin{algorithm}
\caption{MC-SIR algorithm}
\label{alg:MC}
\KwIn{number of initial samples $N$, \newline number of resampled samples $M$}
\KwOut{sampled states $\state_1 ... \state_{M}$}
\For{$i \gets$ \KwTo $N$}
{
    sample $\vdemand'_i \sim \prob(\vdemand)$\;
    solve $\state'_i \gets \modelDS(\vdemand'_i)$\;
    get weight $\omega_i \gets \prob(\vm |\state'_i) \prob(\vdemand'_i)$\;
}

\For{$i \gets$ \KwTo $M$}
{
select $\state_i$ from $\{\state'_j\}_{j = 0}^{N}$ with $p(\state'_j) = \frac{\omega_j}{\sum_k \omega_k}$\;
}
\end{algorithm}
\subsection{MCMC} \label{sec:MCMC}
MC sampling with samples drawn in the space of heat exchanges is very inefficient, as many states have a low likelihood given the measurements. Therefore, we propose using MCMC techniques to draw samples directly from the posterior distribution without needing sample re-weighting \cite{murphy2012machine}. 

The core idea is to construct one or more Markov chains whose distribution converges to the desired posterior distribution. 
As for MC sampling, we can not generate system states directly. We, therefore, run the chains in the space of heat exchanges, in which they converge against $\prob(\vdemand|\vm)$. However, since the mapping \eqref{eq:model-ds} is deterministic, this implicitly defines samples following $\prob(\state|\vm)$ as well.

The MCMC algorithm for probabilistic state estimation is summarised in Algorithm \ref{alg:MCMC}. 
The space of heat exchanges is explored by proposing new candidates $\vdemand'_i$ with a probabilistic Markovian update model, i.e., $\vdemand'_i \sim g(\vdemand_{i-1}) $. 
One intuitive way of constructing such an update model $g(\vdemand_{i-1})$ is to add random noise to the sample, i.e., $\vdemand_{i} = \vdemand_{i-1} + \bm{\epsilon}$ with $\bm{\epsilon} \sim \ND(\mathbf{0}, \mathbf{\Sigma})$. 
Choosing this generation process leads to the Metropolis-Hastings algorithm. 
Different MCMC algorithms improve convergence rates by adapting the proposal process $g(\cdot)$. 
For The details on how this is achieved and how convergence is guaranteed, the reader is kindly referred to the corresponding literature, e.g., \cite{biship2007pattern}.
For this work, it is only essential to emphasise that the proposal process always satisfies the Markov condition, i.e., each new proposal $\vdemand'_i$ only depends on the last sample drawn $\vdemand'_{i-1}$ and not any previous samples. Hence, the samples form a chain and must be calculated one after another. This also means that the samples are only generated during the interference process and can not be calculated beforehand. 

After each proposal step, the position of the chain may be changed depending on the ratio of the un-normalised posterior probability of the current position $\vdemand_{i-1}$, and the proposed update position $\vdemand'_i$,
\begin{align}\label{eq:MCMCposterior_ratio}
    \alpha &= \frac{\prob(\vm |\modelDS(\vdemand'_{i})) \prob(\vdemand'_i)}{\prob(\vm |\modelDS(\vdemand_{i-1})) \prob(\vdemand_{i-1})}.
\end{align}
If $\alpha \geq 1$, the proposal $\vdemand_i$ is accepted; if $\alpha < 1$ the proposal is accepted with probability $\alpha$. If the proposed state is accepted, it is appended to the chain and becomes the new current position. 
Otherwise, the last position of the chain is repeated. 
Thereby, regions of high posterior probability are visited relatively more often than regions of low posterior probability. 
For sufficiently long chains, the samples are guaranteed to resemble samples from the posterior distribution $\prob(\vm |\modelDS(\vdemand)) \prob(\vdemand)$ \cite{biship2007pattern}. 

\begin{algorithm}
\caption{MCMC algorithm}
\label{alg:MCMC}
\KwIn{starting point $\vdemand_0$, number of steps $N$}
\KwOut{sampled demands $\vdemand_1 ... \vdemand_{N}$}
\For{$i \gets$ \KwTo $N$}
{
    generate candidate $\vdemand'_i \gets \mathbf{g}(\vdemand_{i-1})$\;
    solve $\state'_i \gets \modelDS(\vdemand'_i)$\;
    $\alpha \gets \min\left(\frac{\prob(\vm |\state'_i) \prob(\vdemand'_i)}{\prob(\vm |\state_{i-1}) \prob(\vdemand_{i-1})} , 1\right)$\;
    sample from uniform distribution: $\epsilon \gets U[0,1]$\;
    \uIf{$\alpha \geq \epsilon$}
    {
        accept sample $\vdemand_i \gets \vdemand'_i$\;
    }
    \Else
    {
        reject sample $\vdemand_i \gets \vdemand_{i-1}$\;
    }
}
\end{algorithm}

\subsection{Fast MCMC by replacing NR with DNN}\label{sec:MCMC_NN}
MCMC techniques require fewer samples than MC sampling, but the number of samples needed is still in the order of $10^4$ to $10^5$ for our applications. 
Each sample requires solving \eqref{eq:model-ds} to calculate its acceptance probability. 
For the problems we are interested in, obtaining a single NR solution takes up to a few seconds; see also \cite{tol2021development}, which renders these methods computationally infeasible for an online setting.

Therefore, we propose to remove this computational bottleneck by using a DNN to approximate the NR solver. 
I.e., we train a DNN $\NN$ with parameters $\NNweight$ to approximate \eqref{eq:model-ds},
\begin{align}
    \NNstate = \NN (\vdemand),
\end{align}
where $\NNstate$ is the prediction of the DNN for the NR solution.
Neural networks are known to be universal function approximators and can therefore approximate this map arbitrarily well, given sufficient training data and model capacity.

To gather a training data set $\{\vdemand_i,\state_i\}_{i=1}^N$, one can generate $N$ iid samples for the heat exchanges $\vdemand_i$, e.g., by sampling from the prior distribution $p(\vdemand)$, and use the NR solver to obtain the corresponding grid state $\state_i$.
Once the network is trained to sufficient accuracy, we can replace $\modelDS$ with $\NN$ in \eqref{eq:MCMCposterior_ratio}.
While gathering training data and training the neural network can take some time, it has to be done only once. The forward pass of the trained DNN during MCMC is very fast as it only involves a few matrix multiplications.

Additionally, replacing the NR solver with a DNN enables the use of the Hamiltonian MCMC approach \cite{neal2011mcmc}.
The idea of Hamiltonian MCMC is to include a momentum term in the Markovian update model $g(\cdot)$ to increase the number of samples from high density regions of the posterior.
The momentum calculation requires evaluating the derivative of the logarithm of the posterior with respect to the heat exchanges $\vdemand$, which includes the derivative of $\modelDS(\vdemand)$.
The exact derivative of $\modelDS(\vdemand)$ can be derived using implicit differentiation \cite{hoehre_math_fuer_ingeneure}.
However, this is computationally too costly to be used in a MCMC setting due to the high number of samples.
Backpropagating gradients through the DNN, on the other hand, is computationally cheap.

\section{Implementation} \label{sec:implementation}

We implemented our approach in Python using the TensorFlow package \cite{tensorflow2015-whitepaper}.
The NR and the Hamiltonian MCMC algorithm can then use the built-in automatic differentiation tools.
The code is accessible on GitHub \url{https://github.com/EINS-TUDa/DNN_MCMC4DH}

Since some state variables are exactly determined through equations \eqref{eq:fix1}, and \eqref{eq:fix2}, \eqref{eq:fix3}, we exclude them from the NR procedure and fix their value as given.
This avoids invertibility issues for the Jacobian of \eqref{eq:state} and reduces the dimension of the state vector as well as the number of equations. The convergence speed of the NR algorithm is known to depend strongly on the initial point. Therefore we use an iterative approach similar to the decomposed hydraulic-thermal method proposed in \cite{LIU20161238} to find an optimal starting point. More details on this presolve step are given in \ref{app:NR_details}.


\subsection{DNN Configuration and Training} \label{sec:NN} 
\begin{figure}
    \scalebox{0.6}{\input{NN_MCMC/plots/NN_fig}}
    \caption{Structure of the used DNN. The number of nodes in the first and last layer are equal to the number of heat exchanges and estimated states in each test case. The hidden layers have 100, 250 and 250 Nodes in both cases.}
    \label{fig:NN_scheme}
\end{figure}
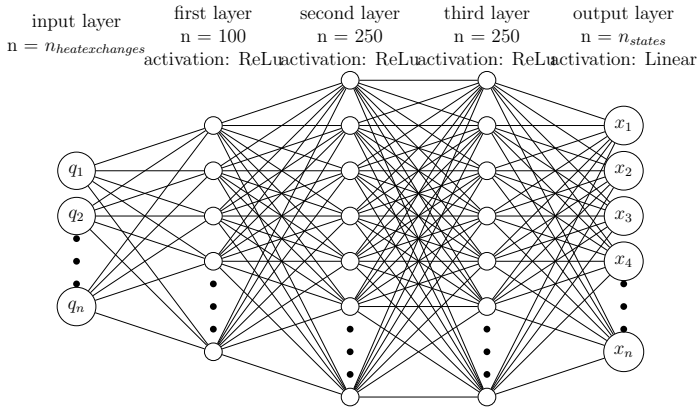

The neural network for our experiments consists of four fully connected layers, as shown in \fig{NN_scheme}. The first three layers have 100, 250, and 250 neurons and ReLU activation functions. 
The fourth layer has a linear activation function, and the number of neurons equals the number of estimated state variables. As for the NR algorithm, we exclude a priori fixed states from the DNN prediction. The input and outputs of the DNN are scaled such that the DNN inputs lie within $[0, 1]$, and the outputs have zero mean and unit variance. All error metrics are calculated after the results are scaled back to their original sizes.


The network is trained to minimise the weighted quadratic difference between the predicted state $\NNstate(\vdemand_i)$ and the true grid state $\state_i$, i.e.,
\begin{align}
\frac{1}{N} \sum_{i=1}^{N} \sum_{\kappa\in \{\temp{} ,\mf{}, \pr{},\tempend{}\}} w_\kappa \|\NNstate^\kappa(\vdemand_i) - \state^\kappa_i\|^2 
\end{align}
where $\state^\kappa$ denotes the dimensions of the state vector encoding temperatures, mass flows, pressures, and line end temperatures, respectively. 
We use values in $kg/s$ for mass flows, $^\circ C$ for temperatures, and $bar$ for pressures
and set the weights as $w_{\mf{}} = 500$ and $w_{\temp{}} = w_{\tempend{}} = w_{\pr{}} = 1$.
These weights are free hyperparameters.
We selected weights that reliably showed good performance for our test case setting without conducting an extensive hyperparameter optimisation. 
Choosing a high weight on the mass flow is motivated by the comparable small numerical values of these state variables given the chosen units. 

The networks for our experiments are trained with 50000 training samples each, using the Adam optimiser \cite{kingma2017adam} with default parameters and a batch size of 32.
The training is terminated when the loss on a validation set of 12500 samples no longer decreases for 20 epochs.
We then use the weights with the lowest validation loss for further calculations.

\subsection{MCMC implementation}
We use the Hamiltonian MCMC implementation provided by the TensorFlow Probability package, which allows a seamless linkage with the neural network \cite{tensorflow2015-whitepaper}.
As far as the parameterisation is concerned, we set the number of leapfrog steps to 1 and determine the step size automatically using a standard TensorFlow optimiser on the first 80\% of the burn-in steps aiming for an acceptance ratio of 75\%.  

As mentioned above, Hamiltonian MCMC requires the backpropagation of the gradients from the un-normalised log-posterior probability to the  distribution from which samples are drawn. 
For our application, this includes calculating the derivative of the likelihood of the measurement with respect to the state, the derivative of the state with respect to the corresponding heat exchange, and the derivative of the log probability of the prior with respect to the heat exchange. 
The former two are straightforward to implement.
The likelihood is given as a normal distribution for which these derivatives are well known.
The derivative of the state with respect to the demands is gathered by standard backpropagation through the DNN. 
Since the heat exchange prior is modelled as a zero-truncated normal distribution, the derivatives are defined piecewise. 
If all heat exchanges are positive, the log-probability of the heat exchange prior is modelled as the log-probability of a non-truncated normal distribution with the same parametrisation as $p(\vdemand)$. The derivative for this case is again well known. 
Otherwise, if at least one heat exchange value is negative, the log-probability is set to negative infinity with a custom-defined gradient containing zeros for all positive entries in the heat exchange vector and +1 for all negative entries. 
This choice of gradient favours positive heat exchanges in the next iteration of the Markov chain and helps prevent it from getting stuck at low demands. 
Since the MCMC algorithm only requires un-normalised probabilities, there is no need to adjust the probability of positive samples for the missing negative probability mass.

\section{Experimental Evaluation}\label{sec:results}
We first demonstrate our algorithm on a slightly modified version of a real-world network which has a classic tree structure ("\RealGrid{}"). 
As a second example, we examine an 18-node grid ("\PTGrid{}"), which includes a ring structure. This topology makes the mass flow direction situation-de-pendent and is, therefore, more challenging for state estimation. 

\subsection{Setup}

\begin{figure*}
     \centering
     \begin{subfigure}[b]{0.6\textwidth}
         \centering
        \includegraphics[width=\linewidth]{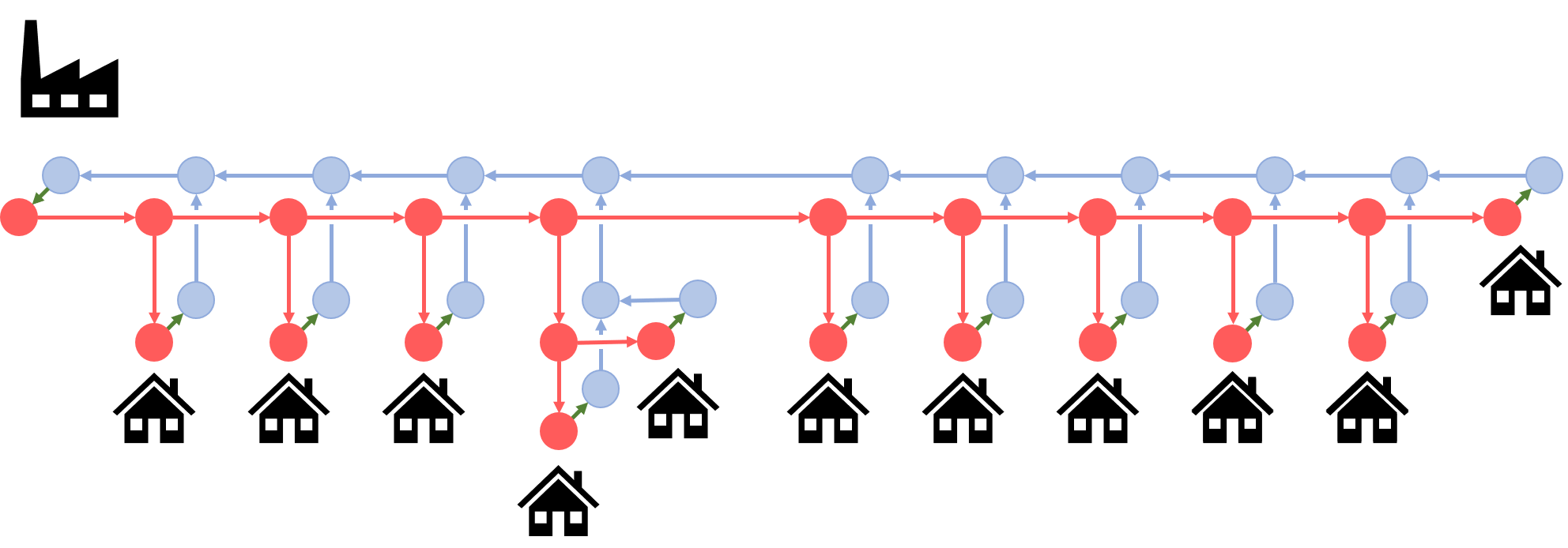}
        \caption{\RealGrid{}}
         \label{fig:grid_real}
     \end{subfigure}
     \hfill
     \begin{subfigure}[b]{0.35\textwidth}
         \centering
    \includegraphics[width=\linewidth]{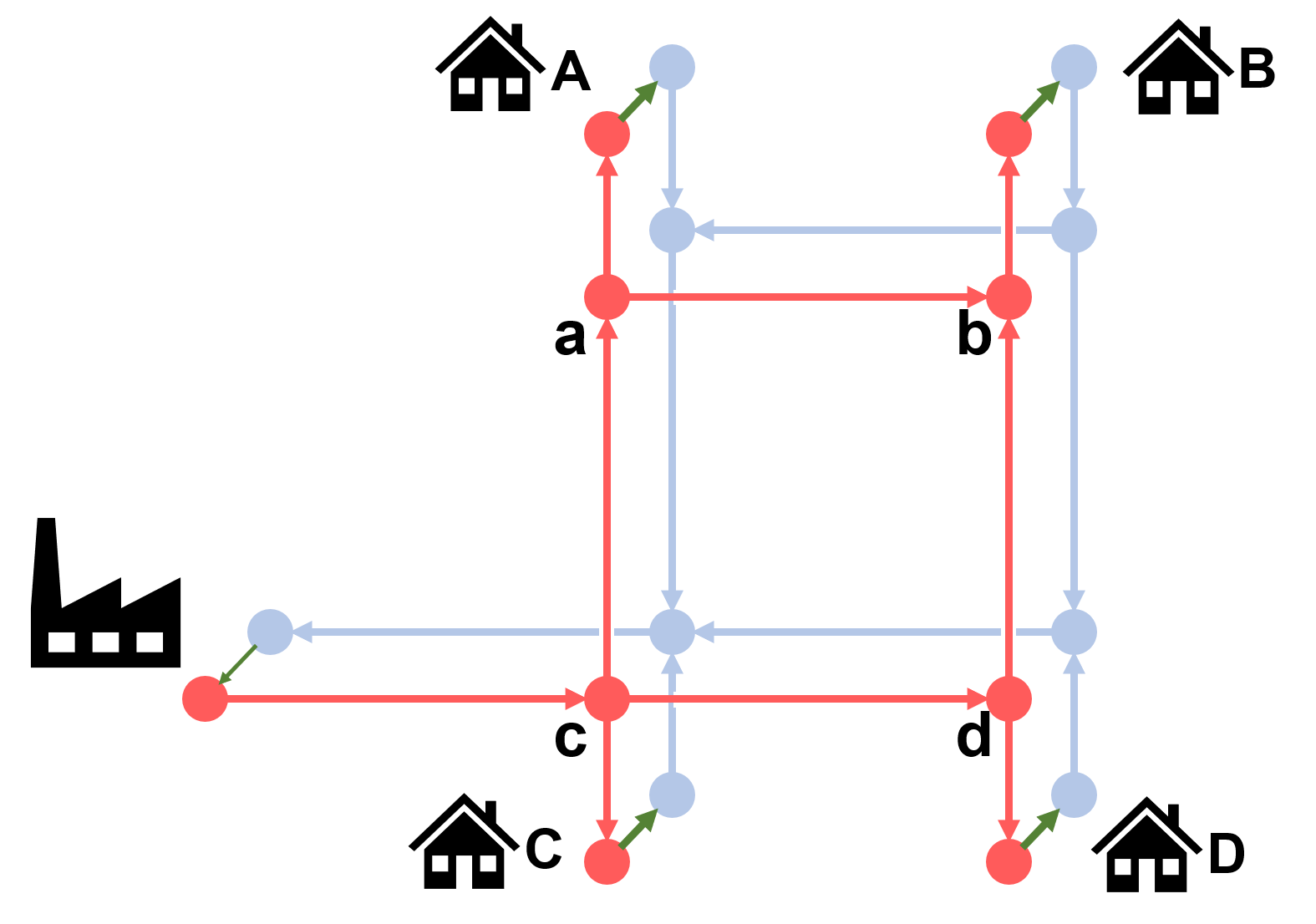}
        \caption{\PTGrid{}}
         \label{fig:grid_Pt}
     \end{subfigure}
     \hfill
        \caption{
        Topology of the two considered heating grid test cases. (a) For a grid with the classical tree structure, the mass flow directions are predetermined. (b) For grids featuring a loop, the mass flow direction is not known in advance. 
        }
        \label{fig:Grids}
\end{figure*}


The first test case \RealGrid{}, shown in \fig{grid_real}, is a modified subsection of an existing district heating grid \cite{BotSte21}.
The grid contains $11$ consumers with a total consumption of $1.7 MW$ and features $57$ edges and $47$ nodes.
It is supplied by one heating station. 
For this grid, historical measurements for each demand in hourly resolution are available, which are used to empirically estimate the prior demand parameters $\demandMean$ and $\demandCov$.
Return temperatures are also derived from these measurements.
As online measurements $\vm$ that we consider for state estimation, we use the return temperature and the mass flow at the heating plant.
We assume that they can be measured with an accuracy of $1\%$.

The heating network \PTGrid{} shown in \fig{grid_Pt} is an artificial example designed to demonstrate the potential of the proposed approach for ring structures which are advantageous in terms of flexibility and supply security.
However, ring structures are challenging for many thermal computation tools since the flow direction in the pipes is not predetermined.
The grid contains 4 demands with a mean total consumption of $0.62\,MW$. The exact parametrisation of the grid and the demand prior can be found in \ref{app:spec_pt}. As in the \RealGrid{} case, we assume measurements for the mass flow $\mf{hp}$ and the return temperature $\temp{r, hp}$ at the heating plant with a standard deviation of $1\%$ of the values for the mean demand condition. 

\subsection{Ground Truth \& Baseline Methods}

The proposed framework (MCMC-DNN) is compared against the Sampling Importance Resampling approach using the NR algorithm (SIR-MC) described in section \ref{sec:MC} as ground truth. 
We decided not to use a NR-based MCMC as baseline as this would require calculating new samples for each measurement, which comes at too high computational costs. 
Even though the SIR-MC requires calculating more samples initially, only the resampling step has to be performed for each new  measurement, which reduces the computational costs overall. 
Consider, however, that this is only true as long as the prior does not change, which might realistically happen in real-life applications. 

\bigskip
Additionally, we compare our approach against linearisation-based probabilistic state estimation (LSE) \cite{BotSte21,MatBotReh21}. 
Here, \eqref{eq:model-ds} is approximated by its first order Taylor expansion around the a priori most probable heat exchange $\demandMean$,
\begin{align} \label{eq:model-lin}
    \state \approx \modelDS(\demandMean) + \Jacobian{\demandMean} \left(\vdemand - \demandMean\right).
\end{align}
$\Jacobian{\demandMean}$ denotes the Jacobian matrix of $\modelDS(\vdemand)$ at $\demandMean$, which is calculated as
\begin{align}
\Jacobian{\demandMean} :=  \left. \dfrac{\partial \modelDS}{\partial \vdemand} \right|_{\vdemand = \demandMean} = 
     - \left[\dfrac{\partial \statequations\big(\state, \demandMean, \setpoints \big)}{\partial \state}\right]^{-1} \dfrac{\partial \statequations\big(\state, \demandMean, \setpoints\big)}{\partial \vdemand}      \label{eq:jacobian}
\end{align}
using the implicit function theorem \cite{hoehre_math_fuer_ingeneure}.
For this baseline, we neglect the truncation of the prior heat exchanges and approximate it with the normal distribution $p'(\vdemand) = \ND \left(\demandMean, \demandCov \right)$. Since the approximated map from heat exchanges to grid states is linear,
the implied  prior distribution of the states is also a Gaussian $\ND\left(\vmu_{\statelin}, \statelinCov\right)$
whose parameters are given by  
\begin{align}
    \vmu_{\statelin} &= 
    \modelDS(\demandMean), \\
    \statelinCov &= 
    \Jacobian{\demandMean} \demandCov \Jacobian{\demandMean}^T. 
\end{align}
As the likelihood is also assumed to be normal, 
standard Bayesian calculus 
states 
that the posterior $p(\state_\statelin|\vm)$ is a Gaussian distribution \\$\ND\left(\vmu_{\statelin}^p, \Sigma_{\statelin}^p \right)$ with parameters
\begin{align}
    \Sigma_{\statelin}^p &= \left(\statelinCov^{-1} + \indexMatrix{M}^T\mCov^{-1}\indexMatrix{M}\right)^{-1}, \label{eq:plin}\\
    \vmu_{\statelin}^p &= \Sigma_{\statelin}^p \left(\indexMatrix{M}^T \mCov^{-1} \vmval + \statelinCov^{-1} \vmu_{\statelin} \right).
\end{align}
Since the number of grid states is always larger than the number of heat exchanges, 
the a priori possible states all lie in a subspace of the full state space.
$\statelinCov$ therefore does not have full rank and is not invertible.
To compute \eqref{eq:plin} 
we add a small noise to the dimensions of $\statelinCov$ that have zero variance, which is done via an eigenvalue decomposition of $\statelinCov$.
All these steps are only performed for the state dimensions which are not fixed by \eqref{eq:fix1}, \eqref{eq:fix2}, \eqref{eq:fix3}.

\subsection{Evaluation metrics}

We evaluate the precision of the DNN predictions via the mean absolute error (MAE) and the mean absolute percentage error (MAPE), where the deviations are individually normalised against the NR values and averaged over a test set consisting of $12 500$ samples which were not used during training. The MAE and MAPE are averaged individually over each dimension of the state vector  $\state^\kappa$ for $\kappa \in \{\temp{}, \mf{}, \pr{}, \tempend{}\}$. 

We also evaluate how well the estimated states $\NNstate$ fulfil the heat grid equations \eqref{eq:state}.
This measure takes the interdependence between the predicted dimensions into account.
I.e., we measure
\begin{align}
    \seLoss = \frac{1}{N} \sum_i
    \|\statequations(\NNstate_i, \vdemand_i, \setpoints)\|^2,
\end{align}
where the average is again taken over the unseen test set.
The computation of $\seLoss$ mixes different physical units, but we have chosen units that render the different numbers comparable in size ($kg/s$ for mass flows, $^\circ C$ for temperatures, and $bar$ for pressures).
This metric is also used as the termination condition for the NR algorithm. 
When comparing the calculation times for the DNN and NR, the termination condition of the NR is adjusted such that both approaches have similar accuracy. 

\bigskip
To evaluate estimated and ground-truth posterior distributions, we use two measures.
First, we capture multivariate effects via the "energy distance" \cite{rizzo2010disco} $\mathcal{E}_{\alpha}(X, Y)$ given as
\begin{equation}
    \mathcal{E}_{\alpha}(X, Y) = 2 \Expval(\|X, Y\|)^{\alpha} -  \Expval(\|X, X'\|)^{\alpha} - \Expval(\|Y, Y'\|)^{\alpha}.
\end{equation}
In our case, the distributions are given via sample sets; thus, $\Expval$ denotes the empirical average. We use  $\alpha = 1$. 
The energy distance approaches $0$ for large sample sets
if and only if the two underlying distributions are equal.
We report the energy distance over all state dimensions as well as
individually calculated for the state dimension $\kappa \in \{\temp{}, \mf{}, \pr{}, \tempend{}\}$ as $\mathcal{E}_{\alpha}^{\kappa}(X^{\kappa},Y^{\kappa})$.

A second measure to compare posterior distributions is motivated by the use case of state estimation for determining safety margins for grid control. 
Reducing supply temperatures in the grid minimises losses, 
but temperatures at customer stations should not violate contractual minimum guarantees. Similarly, reducing supply pressures saves pumping costs but should not lead to insufficient flows.
We, therefore, investigate the accuracy of the $5\%$ quantile of the estimated and true marginal posterior distributions, i.e., we calculate  
\begin{equation}
    \Delta q5^i = |  q_5(X) - q_5(Y) |
\end{equation}
where $q_5(X)$, $q_5(Y)$ are the 5$\%$ quantiles of the $X$ and $Y$ distributions, respectively.
Again, we report this value averaged separately over all  temperatures, pressures and mass flows, as well as the largest values occurring in each state dimension. 
We similarly report the differences of the predicted means as $\Delta m$.

All measures for comparing posterior distributions are computed for 50 measurement values $\vm$ which are computed from heat exchange samples independently drawn from $p(\vdemand)$. All reported numbers are averages of these iterations.

All reported calculation times are derived on a laptop with an Intel i5-8265U CPU processor and 16 GB RAM. 

\subsection{Results Deterministic DNN-Based State Prediction}

\begin{table}[t]
\centering
    \caption{Accuracy of the DNN state predictions for randomly samples heat exchanges}
    \label{tab:performanc_NN}
    \begin{tabular}{l l r r}
        & & \RealGrid{} & \PTGrid{} \\
         \hline
         $\seLoss$ & & 13.48 & 8.56 \\
         \hline
         \parbox[t]{2mm}{\multirow{4}{*}{\rotatebox[origin=c]{90}{MAE}}} & 
            $\temp{} \, [^\circ C]$ & $0.090$ & $0.082$ \\
           &$\mf{} \, [kg/s]$ & $0.744$ & $0.0007$ \\ 
           &$\pr{} \, [mBar]$ & $0.0014$ & $0.029$ \\
           &$\tempend{} \, [^\circ C]$ & $0.089$ & $0.110$ \\
           \hline
         \parbox[t]{2mm}{\multirow{4}{*}{\rotatebox[origin=c]{90}{MAPE $[\%]$}}} &
            $\temp{} $ & 0.10 & 0.11 \\
           &$\mf{} $ &  0.20 & 0.55 \\
           &$\pr{} $ & 0.02 &  0.96 \\
           &$\tempend{} $ &  0.11 & 0.19 
         
    \end{tabular}
\end{table}

\tab{performanc_NN} reports the error margins 
when solving the state equations \eqref{eq:model-ds} via the DNN approach as compared to the exact NR approach.
All MAPE values are below $1\%$, which is deemed practically sufficient. 

\tab{timings} lists the corresponding calculation times.
If the NR algorithm is run until a convergence of $\seLoss \leq 10^{-5}$ is reached, the calculation times are in the order of $1s$ per sample. This is comparable to state of the art solver for heating grids \cite{tol2021development}. 
The computational times for the DNN are more than two orders of magnitude faster than the NR approach, even if the number of NR iterations is reduced to achieve similar accuracy as the DNN solution.
Calculation times can be reduced by another factor of 100
if the evaluation loop is pre-compiled to avoid repeated internal pre-processing of the TensorFlow package. 

Running the complete MCMC-DNN algorithm for 10000 samples, which includes additional effort for proposal generation and derivative computations, is also fast.
The reported time for the LSE approach is determined mainly by the computation of the derivative of the state equations via the implicit function theorem.
The number shows that one such computation is on the order of 10000 times slower than computing an (approximate) derivative via the DNN approach.
MCMC-NR sampling is thus not computationally feasible.
On the other hand, solving the state equations \eqref{eq:model-ds} via the DNN removes this bottleneck of the MCMC algorithm. 

\begin{table}[t]
    \centering
    \caption{Time for calculating 10000 grid states given the heat exchanges. 
    NR timings are reported for different termination conditions. 
    DNN timings are reported using a naive (single pass) and a pre-compiled (graph) evaluation loop. 
    We also report the time needed for 10000 MCMC-DNN steps and the time to calculate the posterior distribution using the LSE model once. 
    }
    \label{tab:timings}
    \begin{tabular}{l c c}
        Time [s]& \PTGrid{} & \RealGrid{} \\
        \hline
        NR ($\seLoss\leq10^{-5}$) & 13402 & 21341 \\
        NR ($\seLoss\leq10^{1}$) & 8511 & 3036 \\
        DNN single pass & 28.8  & 33.9  \\
        DNN graph & 0.19 & 0.42 \\
        \hline
        MCMC-DNN & 11.6  & 12.0 \\
        LSE & 2.6 & 7.3
    \end{tabular}
\end{table}

\subsection{Results Probabilistic State Distributions}
To evaluate the quality of the estimated posterior state distributions, we sample MCMC-DNN results using 10 independent chains with $ 10^{4}$ states each after a  burn-in period of $2 \times 10^{4}$ samples. The results are evaluated against the SIR-MC results as ground truth. 
For the \PTGrid{}, we initially calculate $2\times10^5$ grid states and draw $10^4$ samples during the resampling step. For the \RealGrid{} test case, we calculate $2\times10^6$ samples and draw $10^4$ samples during the resampling step.

\begin{figure}
    \begin{subfigure}{0.4\textwidth}
        \centering
        \includegraphics[width=\textwidth]{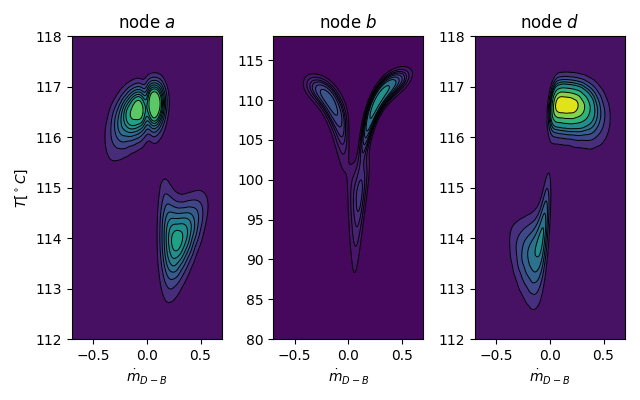}
        \caption{Prior distribution}  
        \label{fig:mfTprior} 
    \end{subfigure}\\
    \begin{subfigure}{0.4\textwidth}
         \centering
         \includegraphics[width=\textwidth]{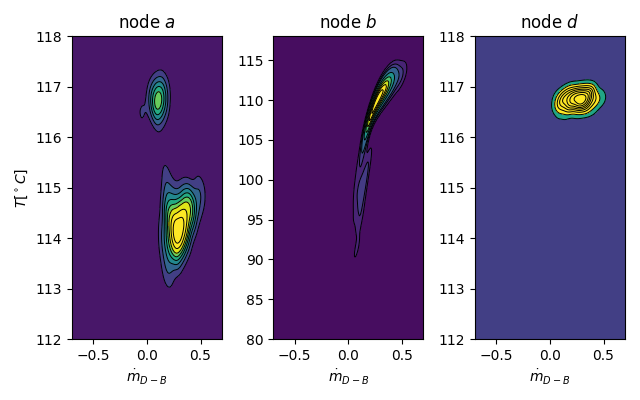}
         \caption{Measurements: $\temp{r,hp} = 49 ^\circ C $; $\mf{hp} = 2.6 kg/s$}
         \label{fig:post_a}
    \end{subfigure}\\
    \begin{subfigure}{0.4\textwidth}
         \centering
         \includegraphics[width=\textwidth]{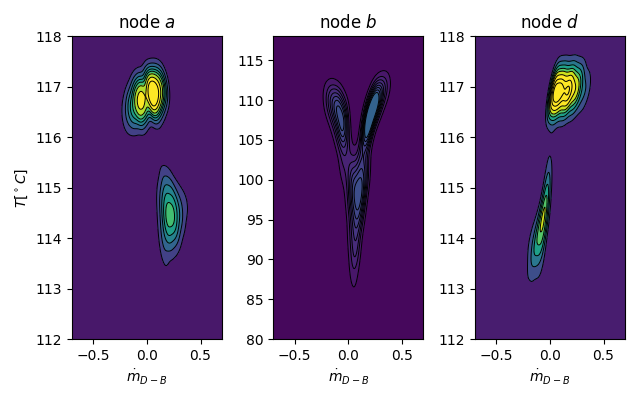}
         \caption{Measurements: $\temp{r,hp} = 48 ^\circ C $; $\mf{hp} = 2.6 kg/s$}
         \label{fig:post_b}
    \end{subfigure}\\
    \begin{subfigure}{0.4\textwidth}
         \centering
         \includegraphics[width=\textwidth]{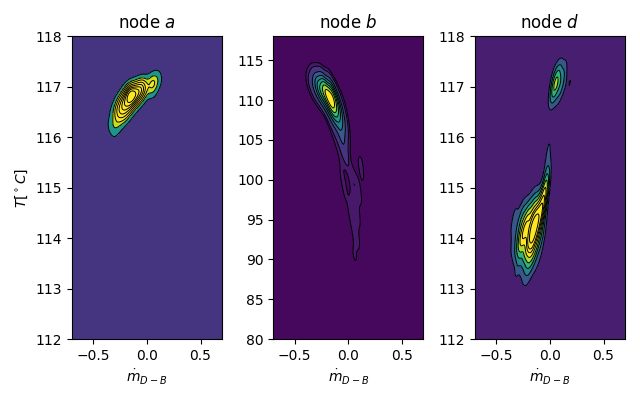}
         \caption{Measurements: $\temp{r,hp} = 47 ^\circ C $; $\mf{hp} = 2.6 kg/s$}
         \label{fig:post_c}
     \end{subfigure}
    \caption{
    Joint distributions of the mass flow $b-d$ and the supply temperatures at $a,b,d$.
    We show the prior and the posterior for different measurements at the heating plant, as given by the SIR-MC ground truth.}
    \label{fig:mfTloop}
\end{figure}

\bigskip

\begin{figure*}[t]
    \centering
    \begin{tabular}{c  c c c}
    & SIR-MC & LSE & MCMC-DNN \\
    \multirow{1}{*}[+6.3em]{\rotatebox[origin=c]{90}{$\Delta p$ at demands}} &
    \includegraphics[width=0.3\linewidth]{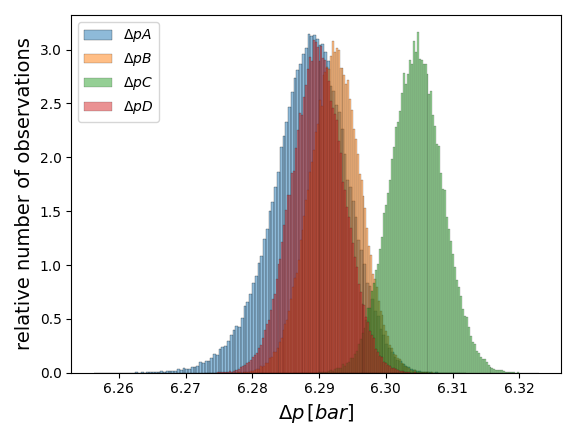} &   \includegraphics[width=0.3\linewidth]{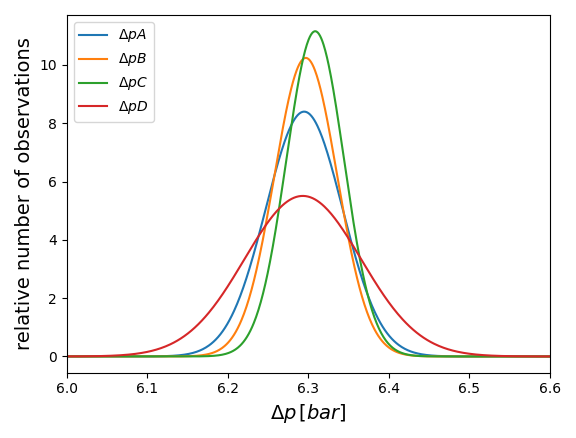} &
    \includegraphics[width=0.3\linewidth]{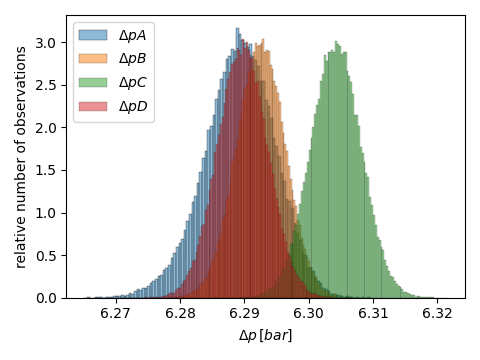}\\
    \multirow{1}{*}[+6.3em]{\rotatebox[origin=c]{90}{$\mf{}$ within loop}} &
    \includegraphics[width=0.3\linewidth]{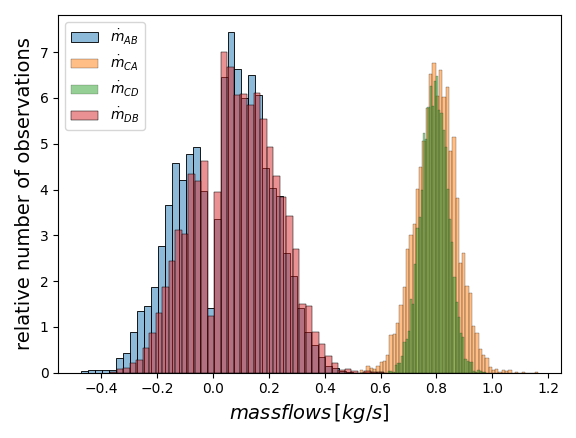} &
    \includegraphics[width=0.3\linewidth]{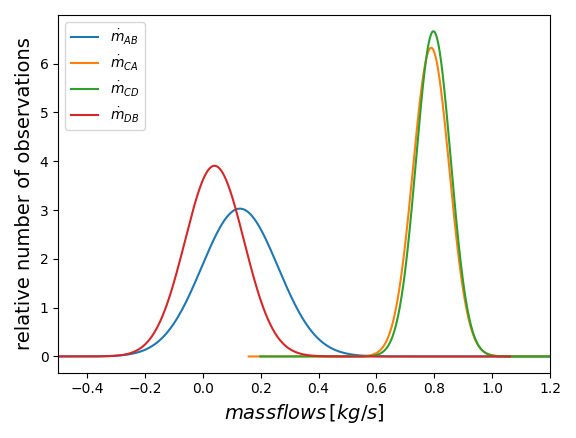}&
    \includegraphics[width=0.3\linewidth]{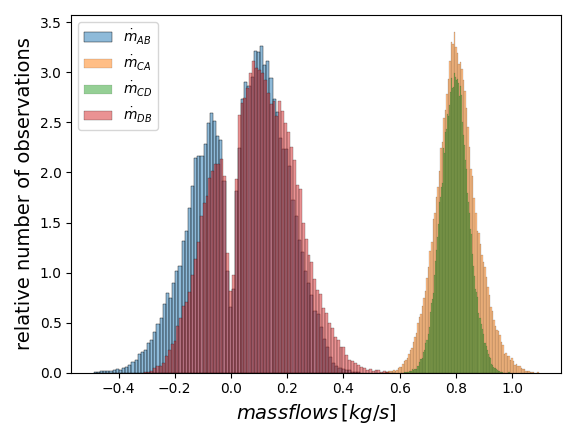} \\
    \multirow{1}{*}[+6.3em]{\rotatebox[origin=c]{90}{$\temp{sup}$ at demands}} &
    \includegraphics[width=0.3\linewidth]{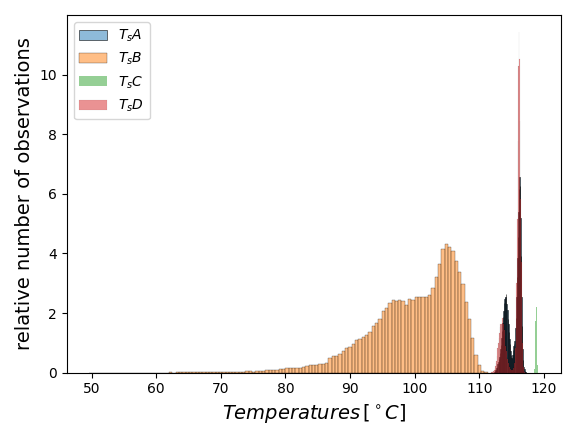}  &
    \includegraphics[width=0.3\linewidth]{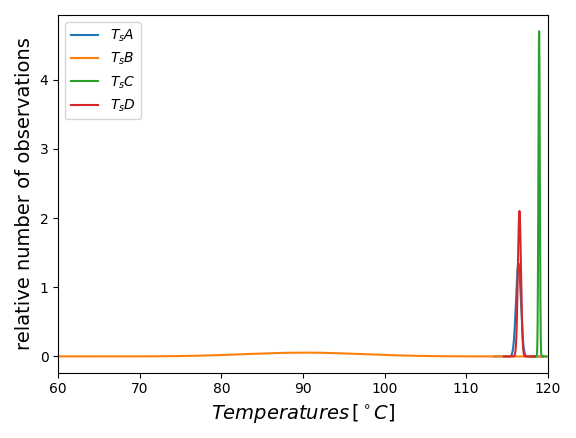} & 
    \includegraphics[width=0.3\linewidth]{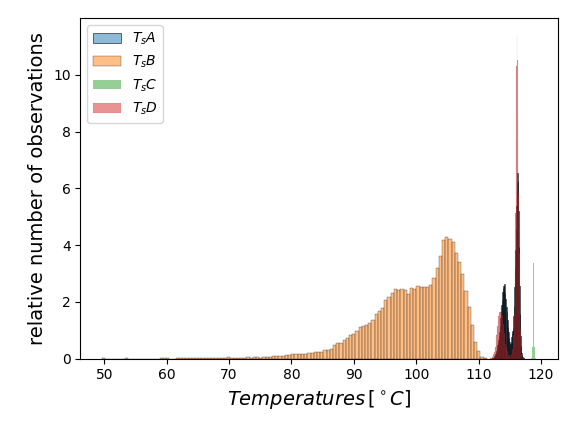}
    
\end{tabular}
    \caption{
    Marginal posterior distributions for exemplary state variables of the \PTGrid{} test case given the measurement values  $\temp{r,hp} = 48.0 ^\circ C$; $\mf{hp} = 2.6 kg/s$. 
    (top row) Pressure differences between the supply and return side of the heat loads, 
    (middle row) mass flows for the ring edges,
    (bottom row) supply temperatures of the heat loads. 
    The left column shows results for the SIR-MC method, considered as ground truth, the middle column for the LSE method, and the right column for the proposed MCMC-DNN approach.
    }
    \label{fig:posteriorsPT}
\end{figure*}
To obtain a first impression of the complexity of the state posteriors, \fig{mfTloop} visualises examples of the dependency structure between state variables for the \PTGrid{} test case via a kernel density estimation plot (KDE-plot). 
Specifically, we show the two-dimensional joint distribution of the mass flow in the loop, i.e., along edge $b-d$, and the supply side temperature at the three potential mixing nodes $a,b,d$. 
The plots show the prior state distribution in \fig{mfTprior} as well as posterior distributions for different measurements; see \fig{post_a} - \fig{post_c}. 

Within the ring structure, the mass flow direction depends on the demands. 
If the demand at A is high compared to the demand at D, the heating fluid flows along $c\rightarrow a$ as well as $c\rightarrow d \rightarrow b \rightarrow a$ and mixes at the top left in node $a$. In this case, the mass flow $\dot{m}_{d-b}$ has a positive sign. 
If the demand at D is higher, the fluid flows along $c \rightarrow d$ as well as $c \rightarrow a \rightarrow b \rightarrow d$. 
In this case, the sign of $\dot{m}_{d-b}$ is negative, and the mixing node is node $d$ at the bottom right. 
Only if the demand difference between A and D is smaller than the demand at B one can observe a positive mass flows for the edges $a \rightarrow b$ and $d \rightarrow b$ simultaneously, which are both very small. Since the heat losses increase with decreasing mass flow, the temperature at the top right node $b$ is reduced in the symmetric case. The asymmetric behaviour between the nodes $a$ and $d$ is due to the different return temperatures of the demands $A$ and $D$.

For all three nodes $a$, $b$, and $d$, the temperature distribution separates into different regimes, depending on the flow $b-d$.
Given the measured values at the heating station, one or more modes of this distribution are likely.
The posteriors are thus clearly non-Gaussian.
Note also that the posterior changes dramatically, even for small changes in the measurements.
These observations prove the demand for a flexible sampling approach 
for state estimation in district heating grids.  
\bigskip

\fig{posteriorsPT} compares the estimated marginal posterior distributions for the grid-loop test case for different state estimation approaches. 
We show exemplary state variables for the measurement $\temp{r, hp} = 47^\circ C, \mf{hp} = 2.6 kg/s$. 
For the pressure differences, one can observe that the assumed SIR-MC ground truth posteriors are unimodal, as are the LSE results.
The LSE method seems to approximate the mean well while overestimating the variances.
In contrast, our proposed MCMC-DNN can approximate the ground truth very well.
The ground-truth posteriors of the mass flows and the supply temperatures, on the other hand, visually do not fall into a common class of distributions and are partly bimodal.
Since the LSE method is bound to always produces unimodal normal posterior distributions, it is not well-suited in this case.
Again, our proposed MCMC-DNN approach yields results that are visually very accurate.
\bigskip

\tab{performance_MCMC_real} list the numerical results for the LSE model and our approach with respect to the SIR-MC baseline for the \PTGrid{} and the \RealGrid{}, respectively. 
Both approaches perform well on the \RealGrid{} test case, featuring a traditional tree structure. The errors for the LSE model tend to be lower, especially for mass flows. The distribution for the states in treelike networks are uni-modal as shown in \cite{BotSte21} and can be approximated sufficiently well using Gaussian distributions. 
Our approach can approximate the distributions as well. However, due to the approximation error of the DNN, the predicted distributions still differ from the ground truth. 

The predictions of the LSE model deteriorate drastically for the \PTGrid{} test case since the marginal distributions are no longer necessarily uni-modal and, therefore, can only be poorly approximated by normal distributions. 
The proposed MCMC-DNN method can approximate arbitrary distributions and thus reaches low error scores for this test case as well. 

\bigskip
As we use a steady-state model to describe the heating grid, the temperatures deviate from the solution of a dynamic model, with the largest deviations occurring after abrupt load changes. In \ref{app:steady_state}, we give an upper bound for these deviations in our \PTGrid{} test case. 
%
For load changes of 30\%, the average supply temperature differences between the two models are below $1.5\, ^\circ C$.
They are thus significantly smaller than the errors from linearising the state equations as done for the LSE approach, reaching up to $14\, ^\circ C$ for $\Delta q5$.

\begin{table*}
    \caption{Numerical indicators of the estimation quality for the state posterior distributions, compared to the MC-SIR baseline. 
    }
    \label{tab:performance_MCMC_real}
    \centering
    \begin{tabular}{l l c c c c}
        &  & \multicolumn{2}{c}{\RealGrid{}} & \multicolumn{2}{c}{\PTGrid{}} \\
        &  & LSE & MCMC NN  & LSE & MCMC NN \\
         \hline
         & combined $\mathcal{E}_{1}$  & 0.86 & 0.54 & 229.2 &  0.196  \\
         \hline
         \parbox[c]{2mm}{\multirow{5}{*}{\rotatebox[origin=c]{90}{$T \, [^\circ C]$}}} &
         $\mathcal{E}_{1}$ & 0.58 & 0.35 &  16.8 & 0.10 \\
         & mean $\Delta q5$ & 0.29 & 0.27 & 3.95 & 0.34 \\
         & max $\Delta q5$ & 2.62 & 2.22  & 14.5 & 2.40 \\
         & mean $\Delta m$ & 0.17 & 0.11 & 2.91 & 0.14 \\
         & max $\Delta m$ & 0.97 & 0.55  & 12.6 & 0.69 \\
         \hline
         \parbox[c]{2mm}{\multirow{5}{*}{\rotatebox[origin=c]{90}{$p \, [mbar]$}}} &
         $\mathcal{E}_{1}$ & 5.19 & 4.45 & 911.6 & 1.31 \\
         & mean $\Delta q5$ & 15.7 & 4.10 & 339.1 & 2.66 \\
         & max $\Delta q5$ & 31.4 & 43.2 & 665.9 & 5.25 \\
         & mean $\Delta m$ & 1.31 & 1.28 & 252.7 & 1.06 \\
         & max $\Delta m$ & 9.41 & 7.61 & 505.0 & 1.60 \\
         \hline
         \parbox[c]{2mm}{\multirow{5}{*}{\rotatebox[origin=c]{90}{$\dot{m} \, [kg/s]$}}} &
         $\mathcal{E}_{1}$ & 0.110 & 0.176  & 9.401 & 0.005  \\
         & mean $\Delta q5$ & 0.010 & 0.098 & 0.055 & 0.009 \\
         & max $\Delta q5$ &  0.054 & 0.358 & 0.151 & 0.028 \\
         & mean $\Delta m$ & 0.007 & 0.066 & 0.043 & 0.009 \\
         & max $\Delta m$ &  0.037 & 0.197 & 0.105 & 0.021 \\
    \end{tabular}
    \vspace{-2mm}
\end{table*}
\section{Conclusion}\label{sec:conclusion}
The paper presents a novel way to combine modern machine-learning methods with a traditional stochastic approach. 
For the little-explored field of probabilistic state estimation in district heating grids, we offer a highly exact approach that is fast enough for online decision-making and whose uncertainty prediction is not limited to any standard distribution class. 
More specifically, given a prior distribution over the heat exchanges and measurements for some of the grid states, the proposed DNN-MCMC algorithm yields samples whose distribution is proportional to the desired posterior distribution over the grid states. 

To this end, we run Markov Chains in the space of heat exchanges and evaluate each sample's probability in the space of grid states against the measurements. 
The mapping from heat exchanges to grid states is encoded using a DNN. 
In our experiments, this reduces the calculation times by a factor of over 50000 compared to a classical solver of the nonlinear grid equations.
This makes the MCMC approach computationally feasible in the first place. 
Additionally, the DNN encoding enables the more efficient Hamiltonian Markov Chain Monte Carlo algorithm to be used as DNNs can be easily differentiated. 

We obtain computation times for the posteriors in the range of tens of seconds. 
This seems reasonable for heating grids where the fluid travelling times in the pipes are typically in the range of multiple minutes. 
While our approach yields good results on traditional tree-like network layouts, it excels most at more complex layouts featuring cycles. 
Unlike state estimation based on Gaussian uncertainties, our approach can closely approximate the non-standard, potentially multi-modal probability distributions observed for the grid states. 
It thus enables tightening the safety margins for temperature and pressure control, that otherwise would have to be chosen very large. This is an important ingredient for more efficient grid operations.

One current limitation of our approach is the restriction to a steady-state analysis. 
We choose this setting to keep the description of the grid equations short, focusing on the probabilistic procedure.
However, the paper's main idea would be well applicable to dynamic grid models as well. 
Taking into account the travel times of temperature waves would further increase the estimation accuracy, especially for large grids and when significant temperature changes occur. 
Extending our idea to the dynamic setting requires more effort w.r.t. the NN network structure, load priors over time, and an efficient training procedure.
This as well as deriving actual control signals from the state estimates is subject to future work.

\section{Acknowledgements}
This research was funded by the German Federal Ministry for Economic Affairs and Climate Action (BMWK) under project number 03EN3012A.
The authors also gratefully acknowledge the computing time provided to them on the high-performance computer Lichtenberg at the NHR Center NHR4CES at TU Darmstadt, which is funded by the Federal Ministry of Education and Research and the state governments participating on the basis of the resolutions of the GWK for national high performance computing at universities.

%% file: NN_MCMC/plots/NN_fig.tex
\centering
\begin{tikzpicture}
\node at (0,7) [circle,draw] (01){$q_1$};
\node at (0,6) [circle,draw] (02){$q_2$};
\node at (0,4) [circle,draw] (03){$q_n$};
\node at (3,8) [circle,draw] (21){};
\node at (3,7) [circle,draw] (22){};
\node at (3,6) [circle,draw] (23){};
\node at (3,5) [circle,draw] (24){};
\node at (3,3) [circle,draw] (25){};
\node at (6,9) [circle,draw] (41){};
\node at (6,8) [circle,draw] (42){};
\node at (6,7) [circle,draw] (43){};
\node at (6,6) [circle,draw] (44){};
\node at (6,5) [circle,draw] (45){};
\node at (6,4) [circle,draw] (46){};
\node at (6,2) [circle,draw] (47){};
\node at (9,9) [circle,draw] (61){};
\node at (9,8) [circle,draw] (62){};
\node at (9,7) [circle,draw] (63){};
\node at (9,6) [circle,draw] (64){};
\node at (9,5) [circle,draw] (65){};
\node at (9,4) [circle,draw] (66){};
\node at (9,2) [circle,draw] (67){};

\node at (12,8) [circle,draw] (81){$x_1$};
\node at (12,7) [circle,draw] (82){$x_2$};
\node at (12,6) [circle,draw] (83){$x_3$};
\node at (12,5) [circle,draw] (84){$x_4$};
\node at (12,3) [circle,draw] (85){$x_n$};

\node at (0, 5.5)[circle,fill,inner sep=1.5pt]{};
\node at (0, 5)[circle,fill,inner sep=1.5pt]{};
\node at (0, 4.5)[circle,fill,inner sep=1.5pt]{};
\node at (3, 4.5)[circle,fill,inner sep=1.5pt]{};
\node at (3, 4)[circle,fill,inner sep=1.5pt]{};
\node at (3, 3.5)[circle,fill,inner sep=1.5pt]{};
\node at (6, 3.5)[circle,fill,inner sep=1.5pt]{};
\node at (6, 3)[circle,fill,inner sep=1.5pt]{};
\node at (6, 2.5)[circle,fill,inner sep=1.5pt]{};
\node at (9, 3.5)[circle,fill,inner sep=1.5pt]{};
\node at (9, 3)[circle,fill,inner sep=1.5pt]{};
\node at (9, 2.5)[circle,fill,inner sep=1.5pt]{};
\node at (12, 4.5)[circle,fill,inner sep=1.5pt]{};
\node at (12, 4)[circle,fill,inner sep=1.5pt]{};
\node at (12, 3.5)[circle,fill,inner sep=1.5pt]{};

\foreach \x in {1,...,3}
    \foreach \y [count=\yi] in {1,...,5}  
      \draw (0\x)--(2\yi);
\foreach \x in {1,...,5}
    \foreach \y [count=\yi] in {1,...,7}  
      \draw (2\x)--(4\yi);
\foreach \x in {1,...,7}
    \foreach \y [count=\yi] in {1,...,7}  
      \draw (4\x)--(6\yi);
\foreach \x in {1,...,7}
    \foreach \y [count=\yi] in {1,...,5}  
      \draw (6\x)--(8\yi);

\node[align=center] at (0,10) {input layer \\ n = $n_{heat exchanges}$};

\node[align=center] at (3,10) {first layer \\ n = 100\\ activation: ReLu};
\node[align=center] at (6,10) {second layer \\ n = 250 \\ activation: ReLu};
\node[align=center] at (9,10) {third layer \\ n = 250\\ activation: ReLu};
\node[align=center] at (12,10) {output layer \\ n = $n_{states}$\\ activation: Linear};
\end{tikzpicture}

%% file: NN_MCMC/10_appendix.tex
\appendix 

\section{Presolving NR algorithm} \label{app:NR_details}
The convergence of the NR algorithnm is known to depend strongly on the initial point. Therefore, a presolver is used to speed up calculations described in Algorithm \ref{alg:NR_presolve}. The algorithm is able to find a find points close to the root of \eqref{eq:state} in a few iterations, but converges slowly towards high precision, therefore complementing the NR algorithm well.

\SetKwInOut{Parameter}{Parameter}
\begin{algorithm}
\caption{NR presolver}
\label{alg:NR_presolve}
\KwIn{demand sample $\vdemand$}
\Parameter{minimal precision $\alpha_1$, minimal gain $\alpha_2$}
\KwOut{starting point $\state$ for NR solver}

$l \gets \infty$\;
$g \gets \infty$\;
separate states $\state^{\kappa}$ in thermal states and hydraulic part\;
$\state^{T} \gets \state^{\kappa}$ with $\kappa \in \{\temp{}, \tempend{}\}$\;
$\state^{h} \gets \state^{\kappa}$ with $\kappa \in  \{\mf{}, \pr{}\}$\;
\While {$l \geq \alpha_1 \ \mathbf{and} \ g \geq \alpha_2$}
{   
    solve hydraulic equations \eqref{eq_mconv}, \eqref{eq:qDemand}, \eqref{eq:fix3}, \eqref{eq:pipeP} for fixed $\state^{T}$\;
    update mass flows and pressures $\state^{h}$\;
    solve thermal equations \eqref{eq:tmix}, \eqref{eq:fix1}, \eqref{eq:fix2}, \eqref{eq_pipeT} for fixed $\state^{h}$\;
    update temperatures $\state^{T}$\;
    
    $l' \gets \statequations \left([\state^{T}, \state^{h}], \vdemand, \setpoints \right) $\;
    $g \gets \left(l - l'\right)/l$\;
    $ l \gets l'$
}
\end{algorithm}


\section{Parameters of the grid model \PTGrid{}}\label{app:spec_pt} 
The grid model \PTGrid{} consists of a ring structure which connects one heating plant with four demands denoted with $A$ to $D$. 
The mean of the demand prior $p(\vdemand)$ is given by \\
${\vdemand_A = \vdemand_C = \vdemand_D = 200\,kW}$ and ${\vdemand_B = 20\,kW}$.
The prior variances are set to ${\sigma_A^2 = \sigma_D^2 = 7000\,kW^2}$,  ${\sigma_C^2=\sigma_B^2 = 100\,kW^2}$.
Between the demands $A$ and $C$ the normalised cross-correlation is set to \\${\rho_{AC}=-0.9}$, for all other demand combinations it is set to zero. 
The return temperatures are fixed as ${\tempend{A} = 50\,^\circ C}$,  \\${\tempend{B}=60\,^\circ C}$, ${\tempend{C}=55\,^\circ C}$, ${ \tempend{D}=40\,^\circ C}$.
At the heating plant, the supply temperature is $\tempend{HP}=120^\circ C$ and the pressures as $6.5 bar$ and $3 bar$ at the supply and return side, respectively.
For all passive pipes $K_{ij} = 0.028\, bar/(kg/s)^2$ and $\lambda_{ij} = 0.2325 \,W/(m K)$. The pipes directly connected to the consumers have the length $l_{iij} = 70\,m$, the pipes at the heating plant and inside the loop have $l_{ij} = 300\, m$.

\section{Steady-State vs. Dynamic Modeling for State Estimation}\label{app:steady_state}
This section analyses the estimation error when using a steady-state model for a dynamic situation. 
We first derive a theoretical bound for the supply temperature errors and then provide experimental results for the \PTGrid{} test case.

\begin{figure}[h]
    \centering
    \includegraphics[width = 0.5 \linewidth]{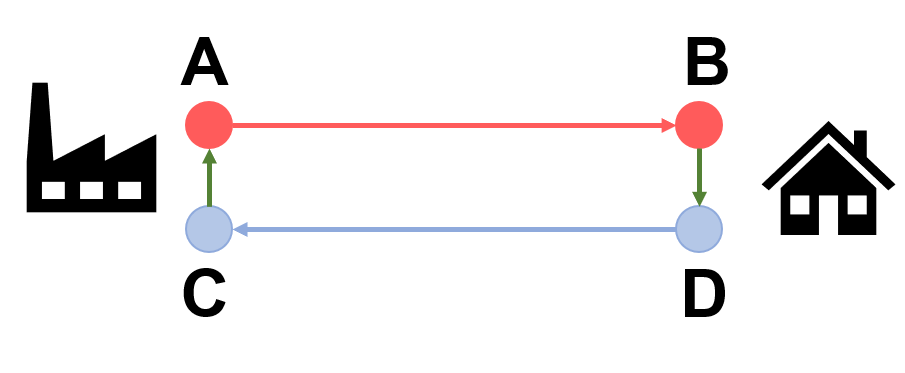}
    \caption{Minimal heating grid for the derivation of the temperature error bounds.}
    \label{fig:Grid} 
\end{figure}

\begin{table*}[ht]
    \centering
    \caption{Upper bound for the error of the supply temperature after a demand shift due to the steady state assumption. The reported results are averages with one standard deviation over 50 random demand samples. 
    }
    \begin{tabular}{l c c}
        scaling factor & 70\% & 130\% \\
        \hline
        Demand A & 
        $(0.296 \pm 0.225) \, ^\circ C$ &  $(0.088 \pm 0.031) \, ^\circ C$ \\
        Demand B & 
        $(1.406 \pm 0.942) \, ^\circ C$ &  $( 0.548 \pm 0.431) \, ^\circ C$ \\
        Demand C & 
        $(0.040 \pm 0.011) \, ^\circ C$ &  $(0.012 \pm 0.003) \, ^\circ C$ \\
        Demand D & 
        $(0.220 \pm 0.082) \, ^\circ C$ &  $(0.071 \pm 0.026) \, ^\circ C$ \\
        \hline
    \end{tabular}
    \label{tab:qss_results}
\end{table*}
Consider a grid consisting only of one heat demand and one heating plant connected by a pair of pipes, as shown in Figure \ref{fig:Grid}.
The dynamic temperature drop along the pipe $(A, B)$ is
\begin{align}
    \temp{B}(t) = (\temp{A}(t-\tau (t)) - \temp{a}) exp \left(- \frac{\lambda_{AB}}{c_p \rho A_{AB}} \tau (t)\right) + \temp{a}, \label{eq:tend_t}
\end{align}
where $\rho$ denotes the density of the heating fluid and $A_{AB}$ is the cross-section area of the pipe \cite{van2017dynamic}. 
$\tau (t)$ denotes the time delay between the fluid entering and leaving the pipe. It can be determined via
\begin{align}
    \int_{t-\tau}^{t} \mf{} (t') dt' = A_{AB} l_{AB} \rho  \label{eq:delay}. 
\end{align}
If the mass flow is constant over time, \eqref{eq:delay} can be solved analytically and replacing the resulting $\tau$ in \eqref{eq:tend_t} yields the steady-state modelling equation~\eqref{eq_pipeT}. 
The demand equation \eqref{eq:qDemand} reads in dynamic form as
\begin{align}\label{eq:demanddyn}
    \demand{} (t) = c_p \mf{} (t) \left(\temp{B} (t) - \temp{D} (t)\right).
\end{align}
Let the system initially be in a steady state with a mass flow $\mf{0}$, time delay $\tau_0$, and temperature $\temp{B,0}$ at the supply side of the demand.

At some time $t_0$, the demand $\demand{}$ is assumed to increase instantaneously, while $\temp{A}$ and $\temp{D}$ remain constant. The mass flow $\mf{}$ then increases instantaneously due to \eqref{eq:demanddyn} to ${\mf{}(t_0) > \mf{0}}$, while ${\temp{B}(t_0) = \temp{B, 0}}$ remains constant at first.
As a consequence, the time delay $\tau$ decreases due to \eqref{eq:delay} and $\temp{B}(t)$ starts to increase slightly due to \eqref{eq:tend_t}. In turn, $\mf{}(t)$ decreases again. 
After some oscillations, the system converges to a new steady state ${\mf{}(t \shortrightarrow \infty)}$ and ${\temp{B}(t \shortrightarrow  \infty)}$ where \\ ${\mf{0} < \mf{}(t \shortrightarrow \infty) < \mf{}(t_0)}$ and
${\temp{B}(t \shortrightarrow  \infty) > \temp{B,0}}$.
Since $\mf{}(t_0)$ is an upper bound for the mass flow during the transition process, solving \eqref{eq:delay} and \eqref{eq:tend_t} with ${\mf{}(t) = \mf{}(t_0)}$ yields an upper bound $\overline{\temp{B}} > \temp{B}(t)$ for the demand's supply temperature during this period.
Load reductions can be analysed analogously.

For the \PTGrid{} test case, we first draw 50 random demand samples from the prior distribution. 
We then scale all demands by 70\% or 130\%, respectively, and calculate the corresponding mass flows at unchanged heating plant supply temperatures. 
Given these mass flows, we calculate the supply temperatures at each demand using \eqref{eq:tend_t} and compare them against the steady-state solution for the scaled demands. 
Table \ref{tab:qss_results} reports the mean and the standard deviation of the computed temperature deviations.
The deviations are significantly smaller than the width of the posterior temperature distributions; see Figure~\ref{fig:posteriorsPT}.
Thus, even though steady-state analysis neglects the time delays in the grid, it provides valuable insights for the grid operators for this test case with pipe lengths in the range of $300\, m$.

The largest deviations between $\overline{\temp{}}$ and $\temp{}(t \shortrightarrow  \infty)$ will occur in large grids or when pipes are not well utilised, i.e., when the mass flow is low with respect to the grid's design conditions. 
In these cases, the cooling effects between the heating plant and the demands are more significant, the temperature differences between the supply and return side of the demands are smaller, and \eqref{eq:demanddyn} is thus more sensitive to supply temperature changes. In the test case, this argumentation can be validated  by observing Demand B, which has a comparatively large deviation. 
Demand B has a lower mean than the other demands but is connected to the grid using the same pipe parameters. Thus, the mass flows are lower in the pipe leading to this demand.